\documentclass[cameraready]{Interspeech}

\title{Merging the Knowledge of LLMs for Automatic Speech Recognition}

\author[affiliation={1}, orcid=0009-0007-9061-5613]{Hayato}{Futami}
\author[affiliation={1}, orcid=0000-0002-2686-2296]{Tatsuya}{Kawahara}

\address{
    $^1$ Graduate School of Informatics, Kyoto University, Japan
}

\email{futami@sap.ist.i.kyoto-u.ac.jp}


\usepackage{comment}
\usepackage{subcaption}
\DeclareMathOperator*{\argmax}{arg\,max}

\begin{document}

\maketitle

\begin{abstract}
Automatic speech recognition (ASR) systems, trained on paired speech-text data, have been improved by leveraging language models (LMs) trained on text-only data.
LM fusion methods such as shallow fusion and density ratio are well-established methods that incorporate external LMs during ASR decoding.
However, they incur additional computational costs due to LM inference, which is particularly problematic for recent larger LMs.
In this study, we propose incorporating external LMs via model merging.
This method integrates the LMs directly into the parameters of an LLM-based ASR model, requiring no additional computational cost at inference.
We formulate domain extension and transfer via arithmetic operations on LoRA parameters.
Experimental evaluations were conducted for the domain adaptation of LLM-based ASR trained on CSJ and LibriSpeech.
We show that our LM merging consistently improved the ASR performance in the target domains, without degrading inference speed or memory footprint.
\end{abstract}

\vspace{-5pt}
\section{Introduction}
End-to-end automatic speech recognition (ASR) models have achieved significant progress in recent years, demonstrating excellent performance in general domains \cite{Radford22-Whisper, Prabhavalkar23-ESR, Peng25-OWSM}.
This success has been driven mainly by large-scale, speech-text paired data coupled with scalable model architectures.
Consequently, these models often suffer from performance degradation in specific domains lacking sufficient paired speech-text data.
As text-only data is much easier to collect, it has been widely explored to apply external language models (LMs) trained on text of the target domains.
Shallow fusion \cite{Kannan17-ALM} is a well-established approach that runs a target-domain LM during ASR decoding, where the log-linear interpolation of the ASR and the LM scores are used.
Density ratio fusion \cite{McDermott19-DR} is an extention of shallow fusion for better domain transfer, where the score of a source-domain LM is further subtracted from the interpolated score.

Recently, following the remarkable success of large language models (LLMs) in natural language processing \cite{OpenAI-GPT4, Dubey24-llama3}, there has been a significant trend toward applying LLMs in the field of ASR.
LLM-based ASR is a prominent application, where pre-trained LLMs are fine-tuned for speech processing tasks by incorporating speech encoders \cite{Wu23-ODA, Fathullah23-PL, Chen24-SALM}.
These models can leverage the knowledge of powerful LLMs as foundations, becoming the new standard approach for ASR \cite{Saon25-Granite}.
Beyond ASR, they are emerging as foundation models capable of solving a wide range of speech processing tasks \cite{Kimi25-KimiAudio, Xu25-Qwen}.
This study also focuses on the LLM-based ASR approach.
LLM-based ASR incorporates the knowledge of an LLM at the initialization phase before ASR training, which cannot be dynamically adjusted according to target domains at inference time.
To this end, LM fusion approaches are applied to adapt LLM-based ASR to the target domain.
Recently, domain-specific LMs are often built by fine-tuning from foundational LLMs.
It is computationally expensive to use such large LMs in shallow fusion or density ratio fusion, where the LMs are called at every decoding step \cite{Hori25-DF, Hu23-MMSF}.

In this study, we propose integrating the knowledge of LLMs into ASR models via model merging.
Model merging is a technique that integrates multiple models by performing arithmetic operations in their weight spaces \cite{Ilharco23-EM, Yang24-MM, Goddard24-MK}.
Our proposed LM merging integrates LMs directly into the ASR model, whereas conventional LM fusion methods rely on off-the-shelf LMs at decoding steps.
A key advantage of our method is that it does not introduce any additional modules or computational overhead over the ASR model.
In this study, we utilize an LLM-based ASR model adapted from a pre-trained LLM using LoRA \cite{Hu22-LoRA}, following \cite{Fathullah23-PL}.
Domain-specific LMs are also bulit via LoRA fine-tuning from the same LLM.
We perform arithmetic operations between the ASR and LM adapters.
We propose two merging methods for domain extension and transfer.
Domain extension merging corresponds to shallow fusion, where the adapter weights of the target-domain LM are added to ASR.
Domain transfer merging corresponds to density ratio fusion, where the adapter weights of the source-domain LM are further subtracted.
These methods are further enhanced by TIES-merging \cite{Yadav23-TIES}, an advanced model merging method that addresses task interference in merging.

We conducted experimental evaluations of ASR domain adaptation in Japanese and English.
We built ASR models based on LLM-jp-3-980M \cite{Aizawa24-LLMjp} and LLaMA3.2-1B \cite{Dubey24-llama3}, trained on CSJ-SPS \cite{maekawa03-CSJ} and LibriSpeech \cite{Panayotov-LS15} corpora, respectively.
The models were then adapted to CSJ-APS and SPGISpeech \cite{ONeill21-SPGI} domains using LMs.
First of all, we confirmed that the proposed LM merging methods consistently improved the ASR performance in the target domains.
We also compared them with conventional LM fusion methods (shallow fusion and density ratio fusion) and $n$-best rescoring \cite{Chen23-LLMR, Ogawa24-ALLMs}.
We observed that while LM merging did not perform as well as LM fusion, it was competitive to $n$-best rescoring on CSJ.
It is important to note that our method does not introduce any additional parameters or latency, unlike LM fusion and rescoring.
We also found that the combination with LM fusion or rescoring further improved the ASR performance.

\vspace{-5pt}
\section{Related work}

\subsection{Language model adaptation for ASR}

As end-to-end ASR models are trained on paired speech-text data, external LMs are often adapted to leverage the knowledge of text-only data.
Shallow fusion is a widely used approach that applies LMs during decoding, in the form of score interpolation between ASR and LM \cite{Kannan17-ALM, Hu23-MMSF}.
Density ratio \cite{McDermott19-DR} extends shallow fusion by subtracting the source-domain LM score to facilitate domain transfer.
Internal LM estimation \cite{Meng20-ILME} insteads subtracts the internal LM score from the end-to-end model.
Recently, delayed fusion \cite{Hori25-DF} has been proposed as an extension of shallow fusion to calculate the LM score with a delay, thereby reducing the number of LLM inference calls.
However, these LM fusion methods suffer from the computational overhead of invoking an LM inference call at each step (or every few steps in \cite{Hori25-DF}).

$N$-best rescoring is also a standard approach that applies LMs as a post-processing step, where $n$-best hypotheses from ASR are rescored using the LM score \cite{Chen23-LLMR, Ogawa24-ALLMs}.
Recently, with the advancement of LLMs, generative error correction has emerged \cite{Yang23-GSREC, Chen23-Hyporadise}.
Given $n$-best hypotheses, LLMs are prompted to generate the correct transcripts.
These post-processing methods require only a single LLM inference call after decoding, which is computationally less expensive than LM fusion.
However, they increase the total parameter footprint.
Moreover, their performance is usually limited for small $n$, especially in the case of greedy decoding ($n=1$), although greedy decoding is preferred in recent LLM-based ASR.

Some studies apply LMs via knowledge distillation \cite{Futami20-DKB, Lee24-OKD}, where the knowledge of a teacher LM is transferred to a student ASR model.
LMs are applied during ASR training, thus the method does not add LM inference cost.
However, since the LMs are only utilized during training and remain static during inference, lacking the flexibility required for domain adaptation.

\subsection{Model merging}
Model merging is a technique that merges the parameters of two or more models to build a new model, fusing their knowledge \cite{Yang24-MM, Goddard24-MK}.
Task arithmetic \cite{Ilharco23-EM} formulates model merging via task vectors, which represent the parameter differences induced by fine-tuning on a specific task.
The study introduces three arithmetic operations: task vector negation, addition, and task analogies.
Task vector addition enables building a multi-task model, and task analogy facilitates improved domain generalization, without requiring access to data or additional training.
Subsequently, several advanced task arithmetic methods have been proposed.
For instance, TIES-merging \cite{Yadav23-TIES} and DARE \cite{Yu23-DARE} sparsify parameters based on their magnitude to mitigate task interference.
Task arithmetic has been further extended to parameter-efficient fine-tuning modules, such as LoRA \cite{Zhang23-CPE, Zhao25-ML}.

While task arithmetic has been applied to a wide range of domains and tasks including LLMs and image recognition \cite{Yang24-MM}, several studies have applied it to ASR.
In \cite{Ramesh24-TVA}, task arithmetic between ASR models has been explored for domain expansion and to simulate training data expansion for low-resource ASR.
In \cite{Su24-S2R}, a synthetic-to-real task vector is used to mitigate the gap of training on synthetic data.
In \cite{Cheng24-TA}, task arithmetic is applied to language expansion for speech translation models.
In \cite{Jing25-RWR}, rare word recognition and translation are tackled via task arithmetic.
That study focuses on merging ASR models trained on synthetic data containing rare words, which differs from our approach of merging LMs into ASR, i.e., cross-modal merging.

\vspace{-5pt}
\section{Language model fusion}
\label{sec:lm-fusion}

Let $\bm{X}$ denote the input acoustic features and $\bm{y}$ denote the text tokens in the transcript.
An end-to-end ASR model estimates $p_{\rm asr}(\bm{y} | \bm{X})$, directly mapping $\bm{X}$ to $\bm{y}$.
LM fusion methods leverage an external LM $p_{\rm lm}(\bm{y})$ during ASR inference by interpolating the output probabilities $p_{\rm asr}(\bm{y} | \bm{X})$ and $p_{\rm lm}(\bm{y})$.

\subsection{Shallow fusion}
\label{sec:shallow-fusion}

Shallow fusion is a widely adopted approach for interpolating $p_{\rm asr}(\bm{y} | \bm{X})$ and $p_{\rm lm}(\bm{y})$ \cite{Kannan17-ALM}.
The end-to-end ASR model searches for the hypothesis that maximizes the log-linear interpolated score of the ASR model and the LM, denoted as:
\begin{align}
\label{eq:shallow-fusion}
\hat{\bm{y}} = \argmax_{\bm{y}}\, [\,\log p_{\rm asr}(\bm{y} | \bm{X}) + \alpha\,\log p_{\rm lm,tgt}(\bm{y})\,].
\end{align}
For domain adaptation, a target-domain LM trained on text data of the target domain, denoted as $p_{\rm lm,tgt}(\bm{y})$, is used.
$\alpha$ in Eq. (\ref{eq:shallow-fusion}) is a hyperparameter.

\subsection{Density ratio}
\label{sec:density-ratio}

Density ratio \cite{McDermott19-DR} is an extension of shallow fusion, which uses Bayes' theorem to define $p_{\rm asr}(\bm{y}|\bm{X})$ using $p_{\rm lm}(\bm{y})$.
The acoustic likelihood for the source and target domains are represented via Bayes' theorem as
\begin{align}
p_{\rm src}(\bm{X} | \bm{y}) = p_{\rm src}(\bm{X}) p_{\rm src}(\bm{y} | \bm{X}) / p_{\rm src}(\bm{y}),
\end{align}
\begin{align}
p_{\rm tgt}(\bm{X} | \bm{y}) = p_{\rm tgt}(\bm{X}) p_{\rm tgt}(\bm{y} | \bm{X}) / p_{\rm tgt}(\bm{y}). 
\end{align}
Give an assumption that $p_{\rm src}(\bm{X} | \bm{y}) = p_{\rm tgt}(\bm{X} | \bm{y})$, the posterior $p_{\rm tgt}(\bm{y} | \bm{X})$ is represented as
\begin{align}
p_{\rm tgt}(\bm{y} | \bm{X}) \propto \frac{p_{\rm tgt}(\bm{y})}{p_{\rm src}(\bm{y})} p_{\rm src}(\bm{y} | \bm{X}). 
\end{align}
We assume $p(\bm{y} | \bm{X})$ and $p(\bm{y})$ are modeled by the end-to-end ASR model and LM, respectively.
The following score is used in beam search:
\begin{align}
\label{eq:density-ratio}
\hat{\bm{y}} = \argmax_{\bm{y}}\, [\,\log p_{\rm asr}(\bm{y} | \bm{X}) + \alpha\,\log p_{\rm lm,tgt}(\bm{y}) \\ \nonumber
- \beta \log p_{\rm lm,src}(\bm{y})\,],
\end{align}
where $p_{\rm lm,src}(\bm{y})$ denotes a source-domain LM trained on the transcripts of the ASR training data.

Both LM fusion methods will improve the ASR performance on the target domain using LMs.
However, they require inference with the target-domain LM (as well as the source-domain LM in density ratio) at every step in the beam search, which increases computational cost during ASR inference.

\vspace{-5pt}
\section{Language model merging}
\label{sec:lm-merging}

We consider an LLM-based ASR model built by attaching a speech encoder to a pre-trained LLM.
The LLM parameters are adapted via the Low-Rank Adaptation (LoRA) for the ASR task, following \cite{Wu23-ODA, Fathullah23-PL, Chen24-SALM}.
For each adapted layer $l$, the parameters $\theta^{(l)}$ are represented as:
\begin{align}
\theta^{(l)}_{\rm asr,src} = \theta^{(l)}_{\rm pre} + \theta^{(l)}_{\rm \Delta asr,src},
\end{align}
where $\theta^{(l)}_{\rm pre}$ denotes the parameters of the base LLM and $\theta^{(l)}_{\rm \Delta asr,src}$ denotes the LoRA adapter parameters for ASR.
Similarly, a target-domain LM is fine-tuned from the same pre-trained LLM, denoted as:
\begin{align}
\theta^{(l)}_{\rm lm,tgt} = \theta^{(l)}_{\rm pre} + \theta^{(l)}_{\rm \Delta lm,tgt},
\end{align}
where $\theta^{(l)}_{\rm \Delta lm,tgt}$ denotes the LoRA adapter parameters for the target-domain LM.

\begin{figure}
  \centering
  \begin{subfigure}{0.43\columnwidth}
    \centering
    \includegraphics[width=\textwidth]{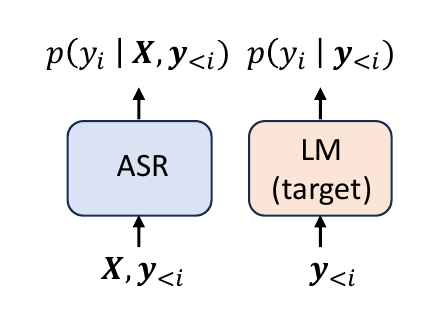}
    \caption{Conventional LM fusion.}
    \label{fig:lm-merging-sub1}
  \end{subfigure}
  \hfill
  \begin{subfigure}{0.55\columnwidth}
    \centering
    \includegraphics[width=\textwidth]{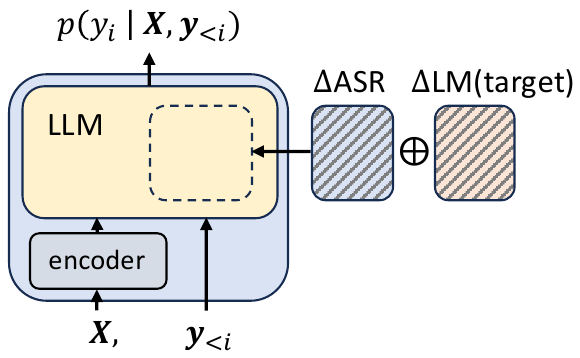}
    \caption{Proposed LM merging.}
    \label{fig:lm-merging-sub2}
  \end{subfigure}
  \caption{Illustration of (a) conventional LM fusion and (b) proposed LM merging for domain extension. LM merging integrates ASR and target-domain LM within the model, which does not add any computational costs at inference.}
  \vspace{-15pt}
  \label{fig:lm-merging}
\end{figure}

\vspace{-5pt}
\subsection{Domain extension merging}
\label{sec:merge-ext}
First, we propose extending the ASR model to the target domain by leveraging the knowledge of the target-domain LM.
This corresponds to shallow fusion described in Section \ref{sec:shallow-fusion}, which integrates the ASR model and the LM at the output log-probability level.
In contrast, we integrate the knowledge of them within the parameter space, as illustrated in Figure \ref{fig:lm-merging}.
We define domain extension merging as:
\begin{align}
\theta^{(l)} = \theta^{(l)}_{\rm pre} + \theta^{(l)}_{\rm \Delta asr,src} + \lambda_{\alpha} \,\theta^{(l)}_{\rm \Delta lm,tgt},
\label{eq:domain-extension}
\end{align}
where the element-wise addition of the parameters is computed with a coefficient hyperparameter $\lambda_{\alpha}$.

Eq. (\ref{eq:domain-extension}) can be further enhanced by advanced merging methods.
Specifically, we consider TIES-merging\footnote{We follow the algorithm \cite{Goddard24-MK} https://github.com/arcee-ai/mergekit} in this study.
First, we trim the parameters to retain the top-$p$\% values by maginitude: 
\begin{align}
\hat{\theta} = {\rm trim}(\theta, p).
\label{eq:ties-trim}
\end{align}
Next, we elect a consensus sign $s^{*}[j]$ for each parameter index:
\begin{align}
s^{*}[j] = {\rm sgn} (\hat{\theta}_{\rm \Delta asr,src}[j] + \lambda_{\alpha}\,\hat{\theta}_{\rm \Delta lm,tgt}[j]).
\label{eq:ties-elect}
\end{align}
Finally, we merge only the parameters whose signs matched the elected signs $s^{*}[j]$:
\begin{align}
\theta[j] = \theta_{\rm pre}[j] + \delta(\hat{\theta}_{\rm \Delta asr,src}, s^{*}[j]) + \lambda_{\alpha}\,\delta(\hat{\theta}_{\rm \Delta lm,tgt}, s^{*}[j]),
\label{eq:ties-merge}
\end{align}
where $\delta(x, s) = x$ if ${\rm sign}(x) = s$ and $0$ otherwise.

This approach is derived from task vector addition in task arithmetic and its extension to parameter-efficient modules \cite{Ilharco23-EM, Zhang23-CPE}.
We build a multi-task model capable of performing both ASR and target-domain LM, by adding the corresponding task vectors $\theta^{(l)}_{\rm \Delta asr,src}$ and $\theta^{(l)}_{\rm \Delta lm,tgt}$\footnote{Task vector is defined as the difference between the task-specific and pre-trained model, corresponding to the LoRA adapter parameters.}.
Unlike \cite{Ilharco23-EM, Zhang23-CPE}, which explore model merging within a single modality, this study explores cross-modal model merging.
We aim to improve target-domain speech-to-text (i.e. ASR) performance without target-domain paired data, by merging a speech-to-text model and a target-domain LM.

\vspace{-5pt}
\subsection{Domain transfer merging}

Secondly, we propose transferring the ASR model to the target domain, which corresponds to density ratio described in Section \ref{sec:density-ratio}.
We define domain transfer merging, by introducing a source-domain LM and its LoRA adapter $\theta^{(l)}_{\rm \Delta lm,src}$, as:
\begin{align}
\theta^{(l)} = \theta^{(l)}_{\rm pre} + \theta^{(l)}_{\rm \Delta asr,src} + \lambda_{\alpha} \,\theta^{(l)}_{\rm \Delta lm,tgt} - \lambda_{\beta}\,\theta^{(l)}_{\rm \Delta lm,src}.
\label{eq:domain-transfer}
\end{align}
Eq. (\ref{eq:domain-transfer}) can be enhanced by TIES-merging, as in Section \ref{sec:merge-ext}.

This approach is derived from task analogies in task arithmetic \cite{Ilharco23-EM, Zhang23-CPE}.
We identify the relationship that ``source-domain LM is to target-domain LM and source-domain ASR is to target-domain ASR''.
The task vector for target-domain ASR can be derived from the remaining three task vectors \cite{Ilharco23-EM}.
Domain transfer merging offers better performance in target domains.
However, it requires access to the source-domain text used for ASR training, which is not always available.
In such cases, only domain extension merging is applicable.

The proposed LM merging offers a clear advantage over the conventional LM fusion explained in Section \ref{sec:lm-fusion}.
While LM fusion requires running one or two LMs in addition to the ASR model during inference, LM merging allows for running only an ASR model into which the parameters of the LM have been merged.
Therefore, it introduces no additional computational overhead at inference time, which is particulary beneficial when using recent large-scale LLMs.
In addition, LM merging retains a single-model forward pass, which can be easily deployed and optimized with existing inference engines (e.g. vLLM).

\vspace{-5pt}
\section{Experimental evaluations}

We conducted experimental evaluations on ASR domain adaptation covering two adaptaion scenarios in Japanese and English.
First, we used the Corpus of Spontaneous Japanese (CSJ) \cite{maekawa03-CSJ}.
CSJ consists of $520$ hours of Japanese oral presentations, including CSJ-APS ($240$h) of academic presentations and CSJ-SPS ($280$h) of simulated public speaking on everyday topics.
In the CSJ experiments, we treated CSJ-SPS as a source domain, where paired data were available to train an ASR model.
Then, CSJ-APS was treated as a target domain, where only text data were available, following \cite{Futami22-NEC}.
In addition to CSJ, we used LibriSpeech \cite{Panayotov-LS15} and SPGISpeech \cite{ONeill21-SPGI}.
LibriSpeech \cite{Panayotov-LS15} comprises $960$ hours of English public-domain book readings.
SPGISpeech \cite{ONeill21-SPGI} is based on corporate earnings calls, representing the financial domain.
In the experiments, we tested domain adaptation from LibriSpeech to SPGISpeech\footnote{We used randomly selected $10$k of $39$k samples for evaluation.}, where only text data were used for SPGISpeech, following \cite{Li23-PLLM}.

We built an LLM-based ASR model by fine-tuning LLM-jp-3-980M \cite{Aizawa24-LLMjp} on CSJ-SPS.
We added a Conformer-based \cite{Gulati20-CF} speech encoder to the LLM, comprising $512$ dimensions, $8$ attention heads, and $12$ layers.
The encoder parameters were fully fine-tuned, while the LLM parameters were updated via LoRA \cite{Hu22-LoRA}, following \cite{Fathullah23-PL}.
We applied LoRA to the key, query, value and output layers of the self-attention modules with a rank $R=8$ and $\alpha=16$.
The model was trained using Adam optimizer of max learning rate $0.0005$ with $25$k warmup steps then decay.
We applied SpecAugment \cite{Park19-SA} as well as speed perturbation \cite{Ko15-AA} of $\times 0.9$ and $\times 1.1$.
We used an auxiliary CTC loss of weight $0.3$.
Decoding was performed using beam search of a beam size $5$ and with joint CTC decoding of weight $0.3$ \cite{Hori17-JCTC}.
We built source-domain and target-domain LMs by LoRA fine-tuning from LLM-jp-3-980M, on CSJ-APS and CSJ-SPS, respecitively.
We also applied LoRA with the same configuration as ASR, trained with Adam optimizer of learning rate $0.00001$ with decay scheduling \footnote{The mean absolute change in the parameters was $0.0001$, while it was $0.0037$ for ASR.}.
The implementation are based on ESPnet \cite{Watanabe18-ES} for ASR with Huggingface's Transformers \cite{Wolf19-HF}.

Table \ref{tab:csj-main} shows the results of our proposed LM merging on CSJ.
We applied domain extension merging (Merge-E) as defined in Eq. (\ref{eq:domain-extension}) and domain transfer merging (Merge-T) in Eq. (\ref{eq:domain-transfer}).
In addition to simple arithmetic operations (i.e., element-wise addition or subtraction), we further applied TIES-merging \cite{Yadav23-TIES} (TIESmerge-T/E) as defined in Eq. (\ref{eq:ties-merge}) with $p=50$\%.
We found that the proposed LM merging methods improved the ASR performance over the baseline in the target domain (eval1).
Domain transfer merging outperformed domain extension merging.
For the source domain (eval3), domain extension merging maintained baseline performance, while domain transfer merging degraded the performance.
These results align with the objective of our methods for domain extension and transfer.
Also, we observed that TIES-merging yielded further gains.

Table \ref{tab:csj-lm-fusion} presents a comparison and combination of our LM merging with other LM adaption methods.
We compared TIESmerge-T with shallow fusion (SF), density ratio (DR), and $5$-best rescoring (Rescore).
We also considered rescoring using the density ratio (Rescore-DR), where the score of the source-domain LM is subtracted as in Eq. (\ref{eq:density-ratio}).
We report the total number of model parameters (Params) and the real time factor (RTF) on an NVIDIA RTX A6000 GPU.
We found that LM merging achieved a CER almost competitive with rescoring.
However, LM merging did not perform as well as SF and DR.
This can be explained by that SF and DR explicitly incorporate the target LM knowledge at every decoding step, while LM merging does not.
Also, we observed a cross-modal alignment problem: increasing the LM coefficient ($\lambda_{\alpha}$ in Eq. (\ref{eq:domain-extension})) leads to a collapse in ASR, making it difficult to sufficiently transfer the LM knowledge.
Note that LM merging has an advantage in model size, RTF and deployment flexibility.
We investigated the combination of LM merging with other methods, where we found that these combinations yielded further performance gains.

\begingroup
\renewcommand{\arraystretch}{1.1}
\begin{table}[t]
  \caption{LM merging for domain adaptation from CSJ-SPS (eval3) to CSJ-APS (eval1).}
  \label{tab:csj-main}
  \centering
  \begin{tabular}{lcccc} \hline
     & \multicolumn{2}{c}{CER(\%) $\downarrow$} \\
     & eval1 & eval3 \\ \hline
    ASR & $13.9$ & $\bm{4.8}$ \\
    +Merge-E & $13.6$ & $\bm{4.8}$ \\
    +TIESmerge-E & $13.4$ & $\bm{4.8}$ \\
    +Merge-T & $\bm{13.3}$ & $5.1$ \\
    +TIESmerge-T & $\bm{13.3}$ & $4.9$ \\ \hline
 \end{tabular}
 \vspace{-10pt}
\end{table}
\endgroup

\begingroup
\renewcommand{\arraystretch}{1.1}
\begin{table}[t]
  \caption{Comparison and combination with conventional LM adaptation methods on CSJ.}
  \label{tab:csj-lm-fusion}
  \centering
  \begin{tabular}{lcccc} \hline
     & CER$\downarrow$ & Params$\downarrow$ & RTF$\downarrow$ \\ \hline
    ASR & $13.9$ & $1.1$B & $0.45$ \\
    +TIESmerge-T & $13.3$ & $1.1$B & $0.45$ \\
    +SF & $12.8$ & $2.1$B & $0.52$ \\
    +DR & $12.5$ & $3.1$B & $0.57$ \\
    +Rescore & $13.3$ & $2.1$B & $0.45$ \\
    +Rescore (DR) & $13.2$ & $3.1$B & $0.46$ \\ \hline
    +TIESmerge-T+SF & $\bm{12.4}$ & $2.1$B & $0.53$ \\
    +TIESmerge-T+DR & $\bm{12.4}$ & $3.1$B & $0.58$ \\
    +TIESmerge-T+Resc & $12.7$ & $2.1$B & $0.45$ \\ 
    +TIESmerge-T+Resc(DR) & $12.8$ & $3.1$B & $0.46$ \\ \hline
 \end{tabular}
 \vspace{-10pt}
\end{table}
\endgroup

Table \ref{tab:libri-main} presents the results of ASR domain adaptation from LibriSpeech to SPGISpeech.
We built an LLM-based ASR model using LLaMA3.2-1B \cite{Dubey24-llama3}.
Different from CSJ experiments, we utlized the WavLM-base-plus SSL model \cite{Chen21-WavLM} as a feature extractor frontend, followed by an E-Branchformer \cite{Kim22-EB} encoder with $256$ dimensions, $4$ attention heads, and $12$ layers.
Both the E-Branchformer encoder parameters and the LoRA adapters in the LLM were updated during training.
As shown in Table \ref{tab:libri-main}, we observed that the proposed LM merging also improved the ASR performance in the target domain (SPGISpeech).
Although the WER reduction was limited compared to other methods, the proposed LM merging offers the advantages of not increasing parameters and RTF over the baseline ASR.
Furthermore, we demonstrated that combining LM merging with other methods yields additional performance gains.

\begin{figure}
  \centering
  \begin{subfigure}{0.45\columnwidth}
    \centering
    \includegraphics[width=0.9\textwidth]{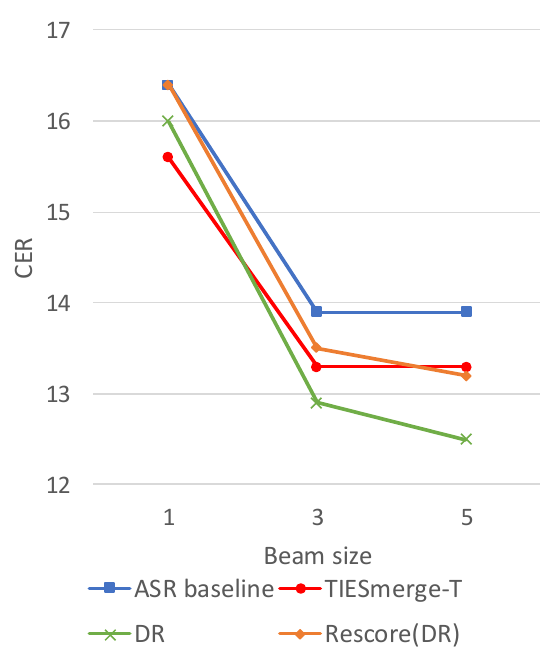}
    \caption{CSJ-SPS to CSJ-APS}
    \label{fig:beam-comparison-sub1}
  \end{subfigure}
  \hfill
  \begin{subfigure}{0.45\columnwidth}
    \centering
    \includegraphics[width=0.9\textwidth]{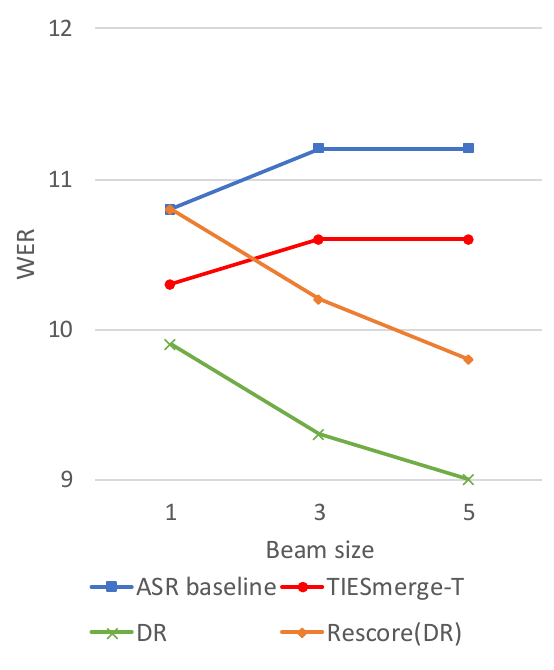}
    \caption{LibriSpeech to SPGISpeech}
    \label{fig:beam-comparison-sub2}
  \end{subfigure}
  \caption{LM adaptation methods on different beam sizes.}
  \label{fig:beam-comparison}
  \vspace{-15pt}
\end{figure}

Figure \ref{fig:beam-comparison} compares TIES-merging (TIESmerge-T) against DR and DR-based rescoring with different beam sizes.
Our LM merging improved ASR even for greedy decoding (beam size of $1$), where rescoring cannot be applied.
The improvements obtained by DR are dependent on the beam size; the gains at beam sizes $1$ and $3$ were not as significant as beam size $5$.
For greedy decoding on CSJ, TIES-merging even outperformed DR.

\begingroup
\renewcommand{\arraystretch}{1.1}
\begin{table}[t]
  \caption{LM merging for LibriSpeech-to-SPGISpeech adaptation.}
  \label{tab:libri-main}
  \centering
  \begin{tabular}{lcccc} \hline
     & WER$\downarrow$ & Params$\downarrow$ & RTF$\downarrow$ \\ \hline
    ASR & $11.2$ & $1.4$B & $0.40$ \\
    +TIESmerge-E & $10.6$ & $1.4$B & $0.40$ \\
    +TIESmerge-T & $10.6$ & $1.4$B & $0.40$ \\
    +SF & $9.1$ & $2.4$B & $0.49$ \\
    +DR & $9.0$ & $3.4$B & $0.56$ \\
    +Rescore & $9.8$ & $2.4$B & $0.41$ \\
    +Rescore(DR) & $9.8$ & $3.4$B & $0.41$ \\ \hline
    +TIESmerge-T+SF & $8.9$ & $2.4$B & $0.49$ \\
    +TIESmerge-T+DR & $\bm{8.8}$ & $3.4$B & $0.55$ \\
    +TIESmerge-T+Resc & $9.3$ & $2.4$B & $0.40$ \\
    +TIESmerge-T+Resc(DR) & $9.3$ & $3.4$B & $0.41$ \\ \hline
 \end{tabular}
 \vspace{-10pt}
\end{table}
\endgroup

\vspace{-5pt}
\section{Conclusions}
\vspace{-5pt}
In this study, we have explored improving the ASR performance in a specific domain, leveraging a target-domain LM.
To this end, we propose integrating LMs into ASR via model merging.
We introduce two merging methods, domain extension and domain transfer, defined by parameter arithmetic between ASR and LMs.
We experimentally demonstrated that LM merging yielded consistent improvement in Japanese and English.
Although the performance improvement was modest compared to SF and DR, LM merging does not increase memory footprint or latency and is executed within a single-model forward pass.
As future work, we will explore more flexible LM merging between different model architectures \cite{Cui26-TM}.

\section{Generative AI Use Disclosure}

Generative AI was used only for proofreading the sentences in the manuscript.

\bibliographystyle{IEEEtran}
\bibliography{mybib}

\end{document}